%% file: main.tex
\documentclass[10pt,twocolumn,letterpaper]{article}

\usepackage{caption}
\usepackage{cvpr}      
\usepackage{amssymb}
\usepackage[accsupp]{axessibility}  

\input{preamble}
\definecolor{cvprblue}{rgb}{0.21,0.49,0.74}
\usepackage[pagebackref,breaklinks,colorlinks,allcolors=cvprblue]{hyperref}
\usepackage{booktabs}
\def\paperID{*****} 
\def\confName{CVPR}
\def\confYear{2026}

\title{TouchMap-OR: Multi-View 3D Mapping of Hand–Surface Contacts}

\author{
Sophokles Ktistakis$^{1}$ \quad
Rui Wang$^{1}$ \quad
Bastian Grande$^{2}$ \quad
Hugo Sax$^{3}$\\[0.5em]
$^{1}$ETH Zurich\\
$^{2}$Institute for Anesthesiology and Perioperative Medicine, University Hospital Zurich\\
$^{3}$Department of Public and Global Health, University of Zurich\\
}
\begin{document}

\twocolumn[{
\renewcommand\twocolumn[1][]{#1}
\maketitle
\begin{center}
    \includegraphics[width=\textwidth]{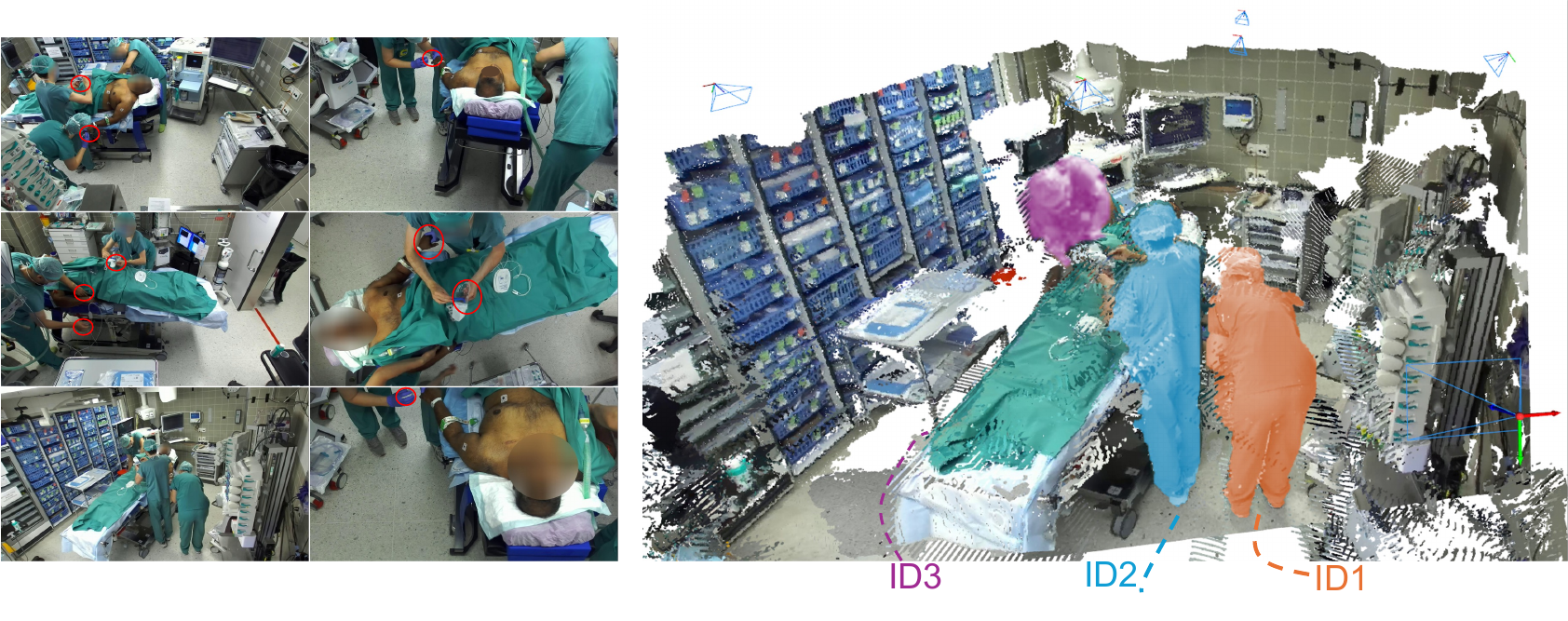}
    {Figure 1. From synchronized multi-view RGB-D observations (left), our system reconstructs a metric 3D representation of the operating room (right) that jointly models clinicians, articulated hands, and the semantic structure of the environment. This representation infers temporally consistent hand–surface contact episodes, identifying which clinician touched which surface, where, and for how long, thereby producing a detailed map of interactions among clinicians, patients, and medical equipment during clinical procedures.}
    \par\vspace{1mm}

\end{center}
}]

\input{sec/0_abstract}

\input{sec/1_intro}
\input{sec/2_relworks}

\input{sec/3_methods}
\input{sec/4_experiments}
\input{sec/5_results}
\input{sec/6_discussion}
\input{sec/7_conclusion}
\input{sec/8_declarations}

{
    \small
    \bibliographystyle{ieeenat_fullname}
    \bibliography{main}
}


\end{document}

%% file: sec/0_abstract.tex
\begin{abstract}
Hand–surface interactions between clinicians, patients, and medical equipment play a central role in pathogen transmission during medical procedures. However, these interactions remain largely unobserved, as current infection-prevention practices rely on manual observation and cannot reconstruct detailed contact histories. In this work we formulate the problem of identity-resolved hand–surface interaction reconstruction in operating rooms and introduce TouchMap-OR, a multi-view RGB-D vision system that models clinicians, articulated hand geometry, and the semantic structure of the clinical environment to infer when and where contacts occur. The system reconstructs globally consistent multi-person 3D skeleton tracks across cameras while estimating articulated MANO hand meshes from RGB observations aligned to depth data. Multi-view hand reconstructions are fused and associated with tracked clinicians to obtain consistent left and right hand trajectories. A semantic 3D model of the operating room is built from multi-view segmentation and depth fusion, enabling reconstructed hand trajectories to be mapped to specific surfaces, including medical equipment, movable objects, and patient body sites. Temporal hand–surface proximity is used to infer contact episodes describing which clinician touched which surface and when. We evaluate TouchMap-OR on recordings from three real anesthesia inductions with manually annotated contact events. TouchMap-OR achieves 0.75 binary contact F1, outperforming tracking-based baselines while maintaining comparable multi-person tracking accuracy and achieving 0.96 identity attribution accuracy.

\end{abstract}

%% file: sec/1_intro.tex
\section{Introduction}
\label{sec:intro}

Preventing pathogen transmission during medical and surgical procedures remains a major challenge. During routine clinical workflows, multiple clinicians interact with shared instruments, monitors, and the patient \cite{schmutz2023my}. These interactions generate dense sequences of hand–surface contacts, creating opportunities for indirect pathogen transmission through contaminated surfaces (fomite transmission) \cite{sax2007my}. Understanding how these contacts unfold in time and space is therefore critical for studying transmission pathways and evaluating infection prevention strategies. However, the dynamics of hand–surface interactions in clinical environments remain poorly characterized. Despite decades of infection prevention research, the current gold standard for assessing hand hygiene compliance remains manual observation, which is labor-intensive, prone to observer bias, and limited in temporal resolution \cite{Jeanes2019}.

Recent work in automated hand hygiene monitoring has explored computer vision as a scalable alternative to manual observation. Systems based on RGB or depth cameras detect dispenser activations or recognize hand hygiene gestures, often achieving accuracy comparable to human observers. Examples include multi-camera systems for detecting hygiene gestures \cite{haque2017towards}, CNN-based recognition of dispenser usage \cite{singh2020automatic}, 3D convolutional models capturing temporal motion patterns during anesthesia induction \cite{Kim2024}, and privacy-preserving approaches using depth-only sensing \cite{chou2018privacy}. While these systems can determine when hand hygiene events occur, they do not observe the contact sequences between hands, surfaces, and patients that govern potential transmission pathways. As a result, they cannot reconstruct contact histories or analyze how contamination may propagate across surfaces and clinical actors.

Outside healthcare, several computer vision approaches have been proposed to estimate hand–object or hand–surface contact. Early work inferred contact using CNN-based hand segmentation \cite{bambach7410583,narasimhaswamy2020detecting}, while more recent methods leverage 3D hand-mesh reconstruction to predict fine-grained contact regions \cite{hasson2019learning,shan2020understanding}. However, these approaches typically assume visible skin, minimal occlusion, and controlled environments. Such assumptions rarely hold in operating rooms, where surgical gloves obscure skin cues and hands are frequently occluded by instruments, clinicians, and medical equipment.

To address this gap, we introduce TouchMap-OR, a multi-view RGB-D vision system that reconstructs identity-resolved hand–surface contact events in real operating rooms. Our approach jointly models three components of the clinical scene: (i) the 3D trajectories of clinicians, (ii) metrically aligned articulated hand meshes, and (iii) a semantic 3D representation of the operating room surfaces. Using synchronized RGB-D cameras, we first estimate multi-person 3D skeleton tracks with stable identities across views. We then reconstruct articulated MANO hand meshes from each camera view and align them metrically to depth observations. These hand reconstructions are fused across cameras and associated with tracked clinicians to obtain consistent left and right hand trajectories for each person. Finally, these reconstructed hand trajectories are integrated with a semantically labeled 3D model of the operating room, obtained by multi-view fusion of segmented RGB-D observations. By evaluating the proximity between hand anchors and semantic surfaces over time, the system infers temporally consistent hand–surface contact episodes that specify which clinician touched which surface, at what location, and for how long. This enables reconstruction of contact histories between clinicians, instruments, and patient surfaces, providing a computational basis for analyzing transmission pathways.

Our key contributions are:

\begin{itemize}[label=--]

\item \textbf{Identity-resolved 3D contact reconstruction in operating rooms.}
We introduce a multi-view RGB-D pipeline that reconstructs hand–surface contact episodes between multiple clinicians and semantically labeled operating room surfaces.

\item \textbf{Multi-view 3D hand reconstruction and person association.}
We fuse MANO-based hand reconstructions across multiple cameras and associate each left and right hand with a persistent clinician identity using multi-view 3D skeleton tracking.

\item \textbf{Semantic 3D contact mapping.}
We map reconstructed hand trajectories to semantically labeled scene surfaces, enabling the reconstruction of temporally consistent hand–surface contact episodes.

\end{itemize}

%% file: sec/2_relworks.tex
\section{Related Works}
\label{sec:methods}

\subsection{Vision-Based Clinical Workflow Monitoring}

Computer vision has increasingly been explored for understanding clinical workflows, with applications including activity recognition, mobility assessment, hygiene monitoring, and operating room scene analysis. Early work focused on recognizing surgical actions and workflow phases from video to support context-aware assistance systems in the operating room~\cite{twinanda2016endonet}. More recent approaches model the operating room through structured representations of actors, objects, and their relationships using semantic scene models and activity graphs~\cite{ozsoy20224d}. Within infection prevention, several works have explored automated hand hygiene monitoring using visual sensing. Multi-camera hospital systems have been proposed to detect hygiene events and track healthcare workers using depth sensing and visual tracking~\cite{haque2017towards}. Other approaches detect alcohol dispenser usage or hygiene gestures using convolutional or spatio-temporal models on RGB or depth data~\cite{singh2020automatic}. Vision-based systems have also been used to assess hand hygiene quality and measure healthcare worker–patient interactions or personal protective equipment compliance~\cite{awwad2019use,chen2016using}. However, these methods primarily detect discrete hygiene events or coarse interactions. In contrast, our work reconstructs \emph{identity-resolved hand–surface contact episodes} across multiple clinicians and objects, enabling detailed analysis of interaction networks and potential contamination pathways in operating rooms.

\subsection{Multi-View Multi-Person Pose Estimation and Tracking}

Human pose estimation has progressed rapidly with deep learning methods for single-view 2D pose estimation, where convolutional networks predict body joint heatmaps from RGB images. Early deep architectures ~\cite{toshev2014deeppose,newell2016stacked} demonstrated the effectiveness of hierarchical feature refinement for human pose prediction, while bottom-up formulations such as OpenPose~\cite{cao2019openpose} enabled real-time multi-person pose estimation by jointly detecting body parts and their associations. Building on these advances, monocular 3D pose estimation methods attempt to recover 3D joint coordinates from single images or videos. Representative approaches include direct 2D-to-3D pose lifting~\cite{martinez2017simple}, volumetric heatmap regression~\cite{pavlakos2017coarse}, and parametric body-model regression frameworks such as HMR~\cite{kanazawa2018end}. While these methods have shown strong performance in controlled settings, single-view approaches remain fundamentally limited by depth ambiguity and frequent identity switches in crowded scenes. To address these limitations, multi-view approaches exploit geometric constraints across synchronized cameras to reconstruct consistent 3D human poses. Early work formulated multi-view pose estimation using pictorial structure models in discretized 3D space~\cite{belagiannis20143d,belagiannis20153d}. More recent methods combine multi-view association with triangulation of 2D keypoints. MVPose~\cite{dong2019fast,dong2021fast} formulates cross-view matching as a multi-way matching problem that groups 2D detections across cameras into consistent person hypotheses before 3D reconstruction. 4D Association~\cite{zhang20204d} extends this paradigm by constructing a spatio-temporal graph over body joints across views and time, jointly solving pose reconstruction and identity tracking. While these approaches focus primarily on accurate 3D pose recovery, our work uses multi-view pose reconstruction as a foundation for stable clinician identity tracking and downstream interaction analysis. 

\subsection{Hand Pose Estimation and Contact Modeling}

Recent advances in deep learning have enabled accurate estimation of hand pose and geometry from RGB images. Early approaches focused on 2D keypoint detection using convolutional networks trained with large-scale datasets and multi-view bootstrapping~\cite{zimmermann2017learning,simon2017hand,iqbal2018hand}. Subsequent work extended these methods to recover full 3D hand pose and shape using parametric hand models such as MANO~\cite{romero2022embodied}. Several methods regress MANO parameters directly from RGB images to reconstruct articulated hand meshes~\cite{boukhayma20193d}, while more recent approaches aim to improve robustness in unconstrained settings through stronger localization and large-scale training data, as demonstrated by WiLoR~\cite{potamias2025wilor}. In parallel, research on hand–object interaction has investigated detecting physical contact between hands and objects. For example, Hasson et al.~\cite{hasson2019learning} jointly reconstruct hands and manipulated objects to infer contact regions, while Shan et al.~\cite{shan2020understanding} introduce large-scale learning of hand–object contact from internet imagery. However, most existing approaches operate in single-view settings and focus on local manipulation tasks with visible hands and limited occlusions. In contrast, our method reconstructs articulated hand meshes independently in multiple views and fuses them into a metric 3D representation aligned with the scene geometry, enabling the detection of identity-resolved hand–surface contact episodes in cluttered multi-person clinical environments.

%% file: sec/3_methods.tex
\section{Methods}
\label{sec:methods}

TouchMap-OR estimates hand–surface contacts (HSCs) of healthcare providers during procedures using multiple calibrated RGB-D cameras. The pipeline outputs (i) globally consistent multi-person tracks with stable IDs, (ii) 3D hand poses per person (left, right), and (iii) contact episodes with surface identity, timestamps, and contact points.

\setcounter{figure}{1}
\begin{figure*}[t]
  \centering
  \includegraphics[width=\textwidth]{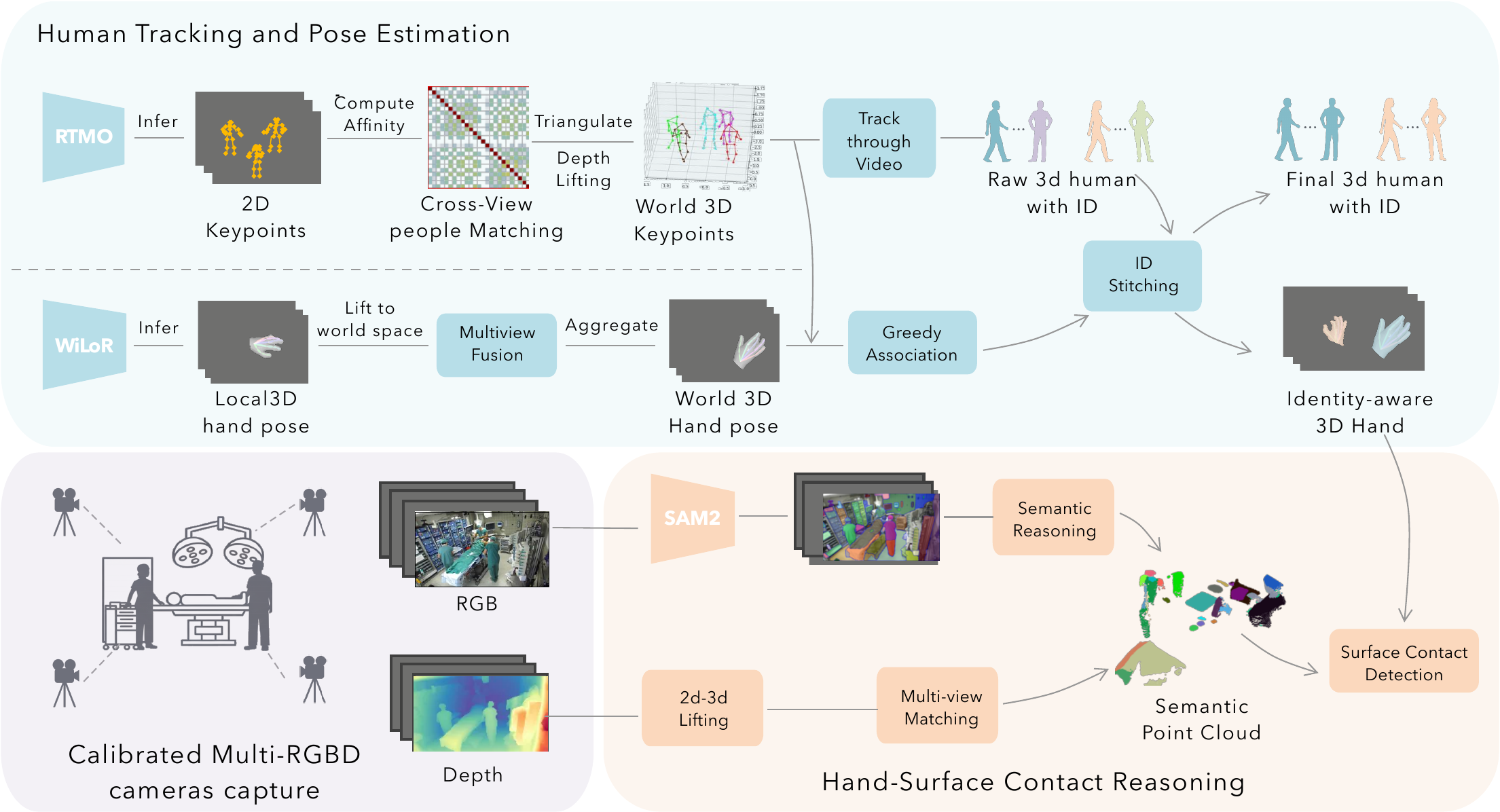}
\caption{Overview of the TouchMap-OR pipeline. Multi-view RGB-D cameras capture the operating room.}  \label{fig:overview}
\end{figure*}

\subsection{Semantic 3D Room Model}\label{subsec2}
To localize HSC events, we construct a semantic 3D model of the operating room. Each of our calibrated RGB-D cameras captures RGB and depth streams, which are first segmented in 2D using SAM2 \cite{ravi2024sam} with sparse, hand-annotated point prompts for key surfaces (e.g., patient, monitors, blanket, tracheal tube, gloves, etc.). Segmented instances are backprojected to 3D, yielding per-camera labeled 3D point clouds. Points from all cameras are fused in the common world frame by voxel down-sampling (10 mm resolution) and majority voting over semantic labels to produce a unified semantic point cloud $S^t$ at frame $t$. The resulting point cloud stores per-vertex surface labels and is indexed by a KD-tree for efficient nearest-surface queries.

\subsection{Per-Camera Detections}
\label{sec:per_camera}

Each RGB-D camera independently estimates body and hand poses of all visible healthcare providers. We use RTMO \cite{lu2024rtmo}, an off-the-shelf multi-person 2D skeleton estimator on each frame. Each joint is denoted $\mathbf{u}_{k}^{(i,t)}=(u_{k},v_{k},1)^{\top}$
with detector confidence $s_{k}^{(i,t)} \in [0,1]$ for joint $k$ in camera $i$ at frame $t$. To recover fine-grained hand geometry, we use WiLoR \cite{potamias2025wilor}, a MANO-based estimator that predicts articulated hand shape and pose from single RGB images. For each detected hand, WiLoR outputs MANO parameters defining a 3D hand mesh
$\mathcal{H}^{(i,t)} \in \mathbb{R}^{778\times3}$ in a wrist-centric frame, along with weak-perspective camera parameters describing only the 2D image projection. To place the mesh in the metric camera coordinate system, we fit the reconstruction to the corresponding depth map via a Sim(3) alignment. Specifically, we sample mean depths in $5\times5$ neighborhoods around projected MANO vertices, back-project valid pixels to 3D, and estimate scale $s$, rotation $R$, and translation $t$ by solving
\[
    \min_{s,R,t}\; \|\, sR\mathbf{x}_{\text{mano}} + t - \mathbf{x}_{\text{depth}} \,\|^2,
\]
using RANSAC to reject occlusions and invalid depth regions. This yields a metrically aligned hand mesh 
$\mathcal{H}_{\text{cam}}^{(i,t)}$, from which we extract hand anchors 
$\mathbf{h}^{(i,t)}$ (palm center and fingertips) and model their reliability 
$\mathbf{\sigma_{\mathrm{fit}}}^{(i,t)}$ using the RMS residual of inlier points after the Sim(3) alignment.

\subsection{Multi-View Association and 3D Person Tracking}
\label{sec:mv_tracking}

To obtain consistent 3D trajectories of all healthcare providers across cameras, we perform multi-view association and joint-wise 3D reconstruction for every frame. Each tracked person $j$ at time $t$ is represented by a set of 26 body joints
\[
\mathcal{X}_j^t=\{(\mathbf{X}_{j,k}^t,\, c_{j,k}^t)\}_{k=1}^{26},
\]
where $\mathbf{X}_{j,k}^t\in\mathbb{R}^3$ is the reconstructed 3D position of joint $k$ and $c_{j,k}^t\in\{0, 1\}$ indicates whether the joint was successfully lifted to 3D via triangulation or depth-lifting, or was unavailable. Each track $\mathcal{X}_j^t$ further maintains a scalar existence score $E_j^t\in[0,1]$ that governs its initialization, confirmation, and removal.

\textbf{Per-camera association.}
For every camera $i$, we project 3D joints of maintained person tracks into the image plane using the known calibration $\pi_i(\cdot)$. Each projected track is compared with all 2D skeleton detections in camera $i$ using the mean per-joint pixel error (MPJPE) over confident joints:
\[
\begin{aligned}
E_{\mathrm{MPJPE}}^{(j,d,i,t)}
&=
\frac{1}{|\mathcal{K}_{i,t}|}
\sum_{k \in \mathcal{K}_{i,t}}
\left\|
\pi_i\!\left(\mathbf{X}_{j,k}^{t^-}\right)
-
\mathbf{u}_k^{(i,t)}
\right\|_2, \\
\mathcal{K}_{i,t}
&=
\left\{
k : s_k^{(i,t)} \ge \tau_{\mathrm{joint}}
\right\}.
\end{aligned}
\]
where $s_k^{(i,t)}$ is the detector confidence and $\tau_{\mathrm{joint}}$ a confidence threshold.
Associations with $E_{\mathrm{MPJPE}}$ above $\tau_{\mathrm{MPJPE}}$ are discarded, and the resulting cost matrix is solved with the Hungarian algorithm to obtain one-to-one matches between existing tracks and detections.
Unmatched detections are kept for cross-view grouping in the birth stage.

\textbf{Triangulation updates.}
For each matched track $j$ and joint $k$, we collect all cameras where that joint is visible and confident.
Epipolar consistency is enforced by rejecting view pairs whose symmetric point–line distance under the fundamental matrix $F_{im}$ exceeds $\tau_{\mathrm{epi}}$.
If at least $V_{\min}$ views remain, the joint is reconstructed by weighted non-linear least squares:
\[
\mathbf{X}_{j,k}^t
=
\arg\min_{\mathbf{X}\in\mathbb{R}^3}
\sum_{i\in\mathcal{V}_{j,k}^t}
w_{k}^{(i,t)}
\big\|
\pi_i(\mathbf{X})-\mathbf{u}_{k}^{(i,t)}
\big\|_2^2,
\]
where $w_{k}^{(i,t)}\!\propto\! s_{k}^{(i,t)}$.
The optimization is initialized from the best-baseline view pair, and the resulting 3D joints are accepted and stored as $\mathbf{X}_{j,k}^t$ with $c_{j,k}^t=1$ if their mean reprojection error falls below $\varepsilon_{\mathrm{tri}}$.

\textbf{Depth lifting.}
Joints that cannot be triangulated are recovered from RGB–D observations.
For each associated camera $i$ and joint candidate $(\mathbf{u}_{k}^{(i,t)},s_{k}^{(i,t)})$, we extract a local depth patch of size $w\times w$ around $\mathbf{u}_{k}^{(i,t)}$ and compute its mean depth $\tilde{z}$ and variance $\sigma_z^2$. Candidates are accepted if
\[
\sigma_z^2 \le \sigma_{\max}^2, \qquad
s_{k}^{(i,t)}\ge\tau_{\mathrm{joint}}.
\]
Accepted pixels are back-projected to 3D as $\mathbf{X}=\mathrm{backproj}_i(\mathbf{u},\tilde{z})$.
To reject spurious joints whose depth values may belong to foreground occluders, background surfaces, or nearby persons we ensure anatomical plausibility by checking bone lengths between neighboring joints $(k,\ell)$ against nominal values $L_{k\ell}$,
\[
\alpha L_{k\ell} \le
\|\mathbf{X}_{k}-\mathbf{X}_{\ell}\|
\le \beta L_{k\ell},
\]
where $\alpha$ and $\beta$ define allowable length ratios.
Based on these bone constraints, joints are grouped into connected kinematic chains, and the longest consistent chain is assigned as $\mathbf{X}_{j,k}^t$. Depth lifting thereby fills missing joints and complements incomplete triangulated skeletons.

\textbf{Births and ID reuse.}
Unmatched detections across cameras are grouped into person hypotheses by evaluating the mean epipolar distance between detections that share confident joints.
Groups observed in at least two cameras are triangulated as above, retaining joints whose reprojection error is below $\varepsilon_{\mathrm{init}}$.
Each valid group initializes a new track with initial existence $E_j^t=E_{\mathrm{init}}$.
If the 3D centroid of a new candidate lies within a spatial distance $r_{\mathrm{reuse}}$ of an existing, unupdated track, its joints are adopted into that track instead of creating a new identity.

\textbf{Track life-cycle and outputs.}
Track existence increases when at least one joint is updated and decays by factor $\lambda$ otherwise.
A track becomes confirmed once $E_j^t \ge E_{\mathrm{on}}$ and is removed when $E_j^t \le E_{\mathrm{off}}$ or after prolonged inactivity.
The tracker outputs per-frame 3D joint positions $\mathbf{X}_{j,k}^t$, joint availability $c_{j,k}^t$, and the corresponding track identity $\mathrm{id}_j$.

\subsection{Hand Fusion and Person Association}
\label{sec:hand_fusion}

After obtaining metrically aligned hand meshes per camera, we fuse them across views and associate each fused hand to its corresponding tracked person. This stage ensures that every left and right hand in the scene is represented by a single, consistent 3D mesh in the world frame and linked to a unique person identity.

\textbf{Cross-view hand fusion.}
For each frame, all metrically aligned MANO hand meshes $\mathcal{H}_{\mathrm{cam}}^{(i,t)}$ from the individual cameras are transformed into the common world frame using the calibrated extrinsics $T_{\mathrm{cw}}^{(i)}$. Each reconstructed hand is represented by a dense vertex set $\mathcal{H}_{\mathrm{w}}^{(i,t)} \in \mathbb{R}^{778\times3}$, a residual $\sigma_{\mathrm{fit}}$ from the preceding Sim(3) alignment, which serves as a confidence proxy, and it's side label (left, right).  

From each mesh, we extract the 3D palm center $\mathbf{c}^{(i,t)}$. All hand detections within the same frame are grouped across cameras using density-based spatial clustering (DBSCAN) on their palm centers. This procedure merges all views that observe the same physical hand. The representative of each cluster is selected as the member with the lowest residual $\sigma_{\mathrm{fit}}$, whose world-space mesh $\widehat{\mathcal{H}}_u^{(t)}$ defines the fused hand instance used in subsequent stages.

\textbf{Greedy person association.}
Each fused hand is then linked to one of the tracked persons from Section~\ref{sec:mv_tracking} through a greedy distance-based association with temporal persistence. For every hand center $\mathbf{c}_u^t$, we compute its distance to the corresponding side-specific body joints (wrist, elbow, shoulder) of every tracked person:
\[
\begin{aligned}
d_{j}^{(\mathrm{side})}
&=
\min_{k\in\mathcal{J}_{\mathrm{side}}}
\|\mathbf{c}_u^t - \mathbf{X}_{j,k}^t\|_2, \\
\mathcal{J}_{\mathrm{side}}
&=
\{\mathrm{wrist},\mathrm{elbow},\mathrm{shoulder}\}.
\end{aligned}
\]
Temporal persistence enforces continuity of hand–person associations across frames by favoring previous pairings whenever the hand remains spatially consistent with its prior person within a predefined motion gate,
$d_{\mathrm{hand,prev}} < d_{\mathrm{gate}} = v_{\max}\,\Delta t + \delta$,
where \(v_{\max}\) denotes the maximum expected hand velocity, \(\Delta t\) the inter-frame interval, and \(\delta\) a spatial slack to account for minor localization noise.

Associations are ranked by joint priority (wrist first) and accepted if $d_j^{(\mathrm{side})} < \tau_{\mathrm{assoc}}$. Each person maintains two \emph{side slots} (left and right), allowing only one hand to be assigned per side. If a newly detected hand provides a better match, i.e. a closer correspondence to the relevant body joints it replaces the previous occupant of that slot. This mechanism ensures geometric consistency between the hand location and the person’s limb configuration. Hands that remain unmatched within the spatial gate initialize new temporal hand tracks but are not associated with any person identity.

\textbf{Person ID stitching.}
To enhance long-term identity consistency, we apply a stitching procedure that merges fragmented person tracks caused by temporary occlusions and missed detections. During hand-person association, each fused hand remembers its most recent person assignment and casts a re-association vote whenever it links to a different person track in later frames. Over time, these votes accumulate into a correspondence matrix between transient and persistent person IDs. At the end of the sequence, a one-to-one mapping is established by selecting the most frequent and mutually consistent ID pairs, thereby merging short, disconnected track fragments into unified trajectories.

\subsection{Hand-Surface Contact Detection}
\label{sec:contact}

To determine when and where each hand makes contact with the surrounding environment, we estimate the proximity of every fused 3D hand to the semantic surfaces described in Section~\ref{subsec2}.  
For each frame $t$, we query the nearest surface point from the semantic point cloud $S^t$ for all hand anchor points $\mathbf{h}_u^t$ (the palm center and finger-tips).  
A contact candidate is registered when the minimum Euclidean distance between any anchor and a surface point falls below a threshold $\tau_{\mathrm{on}}$. To improve temporal stability and suppress false on/off flickering due to sensor noise, we apply two filtering stages.  
First, all 3D hand coordinates are smoothed over time using an exponential moving average (EMA) filter
$\mathbf{h}_u^t = \alpha\,\mathbf{h}_u^t + (1-\alpha)\,\mathbf{h}_u^{t-1}$,
to reduce frame-to-frame jitter.
Second, we enforce a hysteresis rule on the smoothed hand–surface distance signal.  
A contact is activated once the distance drops below $\tau_{\mathrm{on}}$ and remains active until it exceeds a higher release threshold $\tau_{\mathrm{off}} > \tau_{\mathrm{on}}$.  

For each hand $u$ and surface $\ell$, consecutive frames with an active contact state are merged into a single contact episode 
$(t_{\mathrm{start}}, t_{\mathrm{stop}})$, producing a set of temporally consistent contact intervals
\[
\mathcal{C} = 
\left\{
(\mathrm{id}_j,\, \mathrm{side}_u,\, \ell,\, t_{\mathrm{start}},\, t_{\mathrm{stop}},\, \mathbf{p}_{\mathrm{contact}})
\right\},
\]
where $\mathrm{id}_j$ denotes the associated person identity, $\mathrm{side}_u \in \{\mathrm{left},\mathrm{right}\}$ the corresponding hand side, and $\mathbf{p}_{\mathrm{contact}}\in\mathbb{R}^3$ the representative contact point on the surface. 

%% file: sec/4_experiments.tex
\section{Experiments}

\begin{figure*}[t]
  \centering
  \includegraphics[width=\textwidth]{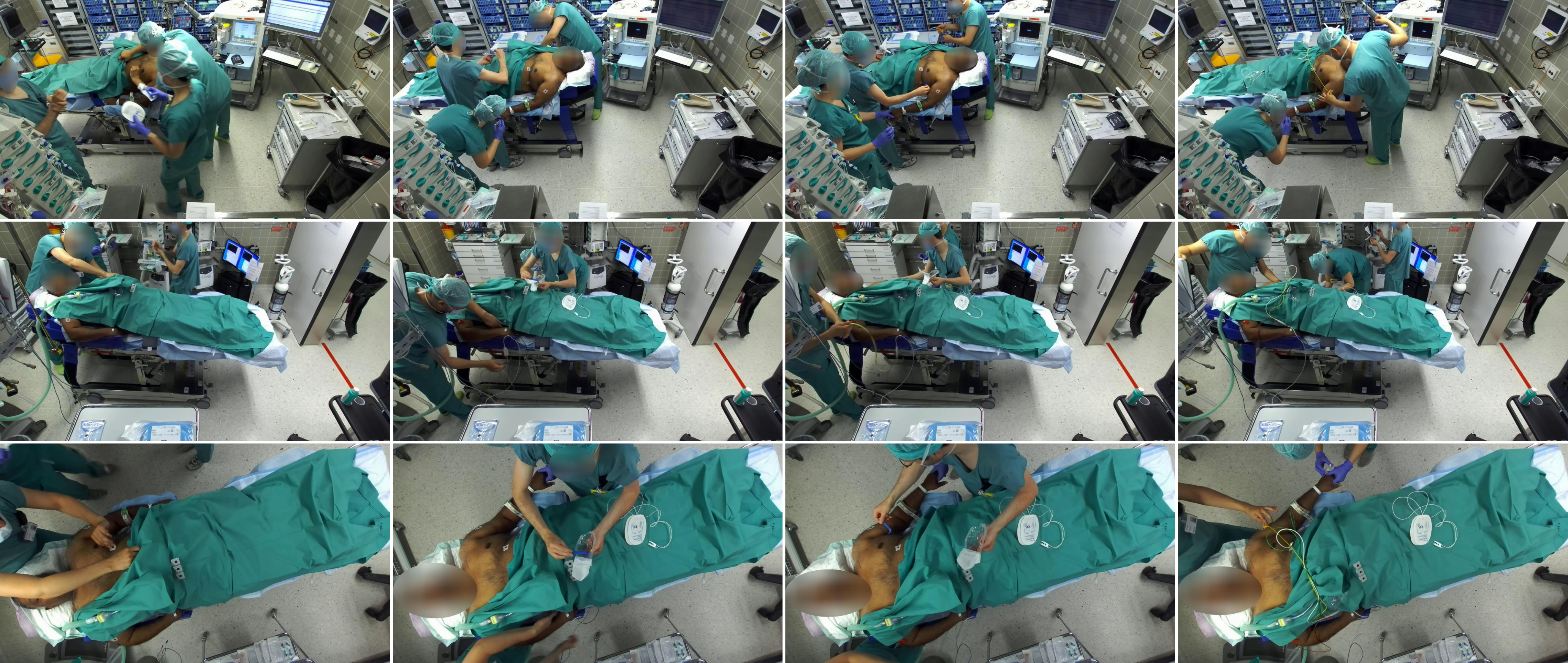}
  \caption{Example sequence from our anesthesia induction dataset. Selected frames from three synchronized camera views (rows) illustrate clinician–patient and clinician–equipment interactions, including intravenous catheter (venflon) placement, patient chest shaving, and ECG cable placement in a cluttered operating-room environment.}
  \label{fig:overview}
\end{figure*}

\subsection{Dataset}
We evaluate our method on a dataset collected during three real anesthesia induction procedures in a clinical operating room. We recorded using six synchronized RGB-D stereo cameras operating at 30 fps and positioned on the walls around the patient. Our recordings span between 5k–6k frames (17k in total), and involve 2–5 healthcare professionals performing routine induction activities such as ECG lead placement, peripheral intravenous catheter (venflon) insertion, patient chest shaving, and preparation of ventilation equipment. The recordings capture both gloved and ungloved hands and include frequent occlusions, overlapping clinicians, and interactions with diverse surfaces, reflecting realistic clinical conditions. The dataset contains 1616 manually annotated hand–surface contact (HSC) episodes spanning interactions with patients, movable medical equipment, deformable objects (e.g., blankets), and fixed environmental surfaces. Each episode is annotated with hand laterality (left/right), surface semantic label, and temporal onset and offset frames. Annotations were jointly produced by the authors in consultation with an anesthesiologist to ensure accurate interpretation of clinical actions and contact events. For identity-aware evaluation, each clinician was assigned a consistent identity label and tracked in 3D. Clinicians were segmented using SAM2, and their identities were associated over time through a semi-automatic tracking pipeline. The resulting tracks were then manually verified to ensure identity consistency. Each contact episode is therefore associated with both the responsible clinician identity and the contact surface, enabling evaluation of clinician–surface interaction attribution.

\subsection{Evaluation Protocol}
We evaluate our system on two aspects: clinician tracking and hand–surface contact detection. For tracking, we report standard multi-person MOT metrics, including IDF1 and the number of identity switches (IDSW). Predicted and ground-truth tracks are associated per frame using Hungarian matching based on the distance between floor-projected person centers with a maximum radius of 0.2 m. Hand–surface contact episodes are defined by clinician identity, hand laterality (left/right), surface label, and temporal interval. Evaluation is restricted to annotated episodes where the hand is visible and a valid semantic surface segmentation is available, excluding frames with full occlusion or missing surface reconstruction. We report three complementary metrics. Episode Recall measures whether a ground-truth contact episode is detected, defined as at least one predicted contact overlapping the ground-truth interval by $\geq{1}$ frame with matching hand laterality and surface label. Contact F1 evaluates framewise binary contact detection independent of surface identity, while Semantic Contact F1 additionally requires the correct surface label. Binary Contact F1 is reported separately because semantic surface attribution can be ambiguous in cluttered scenes or when interacting with small or deformable objects (e.g., gauze or bandage components). For example, when a clinician’s hand is near multiple surfaces, such as a bandaged arm with an inserted venflon and nearby blankets the exact segmentation boundary may not perfectly correspond to the annotated contact surface. Binary Contact F1 therefore evaluates contact detection independent of surface labeling. Finally, Identity Accuracy measures whether predicted contact episodes are attributed to the correct clinician based on the majority ground-truth identity across frames. All metrics are aggregated across recordings and annotated hands.

\subsection{Baselines}

We compare our method against two multi-view 3D pose tracking approaches commonly used for multi-person reconstruction: 4D Association \cite{zhang20204d} and MVPose \cite{dong2019fast,dong2021fast}. Both methods recover identity-consistent 3D skeleton trajectories from synchronized multi-view observations. 4D Association jointly associates 2D detections across views and time to reconstruct temporally consistent 3D poses, while MVPose triangulates multi-view keypoints and performs geometric association to obtain multi-person 3D skeletons. These baselines provide strong multi-view tracking but do not explicitly model hand–object interactions. To evaluate contact detection, we apply the same distance-based contact criterion used in our method to the hand positions produced by the baselines. Hand keypoints from the reconstructed skeletons are compared to the segmented 3D scene, and a contact is predicted when the distance between a hand and the nearest surface falls below a fixed threshold over consecutive frames. The contact is assigned to the closest surface label. This ensures that differences in performance arise from the quality of the reconstructed hand trajectories rather than the contact detection rule.

%% file: sec/5_results.tex
\section{Results}

\begin{table}[t]
\centering
\caption{Multi-person tracking performance.}
\label{tab:tracking_results}
\setlength{\tabcolsep}{6pt}
\begin{tabular}{lcc}
\toprule
Method & IDF1 $\uparrow$ & ID Switches $\downarrow$ \\
\midrule
4D Association & 0.27  & 547 \\
MVPose & \textbf{0.49} & 236 \\
\textbf{TouchMap-OR (Ours)} & 0.48 & \textbf{108} \\
\bottomrule
\end{tabular}
\end{table}

\begin{table*}[t]
\centering
\caption{Evaluation of hand–surface interaction detection and identity attribution. Contacts are detected using a distance threshold of $12\,\mathrm{cm}$, while person identities are matched using a spatial threshold of $0.2\,\mathrm{m}$.}
\label{tab:contact_identity_results}

\setlength{\tabcolsep}{6pt}
\renewcommand{\arraystretch}{1.1}

\begin{tabular}{l c|cc|c|c}
\toprule
 & \textbf{Episode} & \multicolumn{2}{c|}{\textbf{Binary Contact}} & \textbf{Semantic Contact} & \textbf{Identity} \\
\cmidrule(lr){2-2} \cmidrule(lr){3-4} \cmidrule(lr){5-5} \cmidrule(lr){6-6}
Method & Recall & IoU & F1 & F1 & Episode ID Acc. \\
\midrule
4D Association + heuristic & 0.04 & 0.17 & 0.29 & 0.04 & 0.44 \\
MVPose + heuristic & 0.25 & 0.39 & 0.56 & 0.22 & 0.95 \\
\textbf{TouchMap-OR (Ours)} & \textbf{0.54} & \textbf{0.60} & \textbf{0.75} & \textbf{0.37} & \textbf{0.96} \\
\bottomrule
\end{tabular}
\end{table*}

\begin{figure}[t]
  \centering
  \includegraphics[width=\columnwidth]{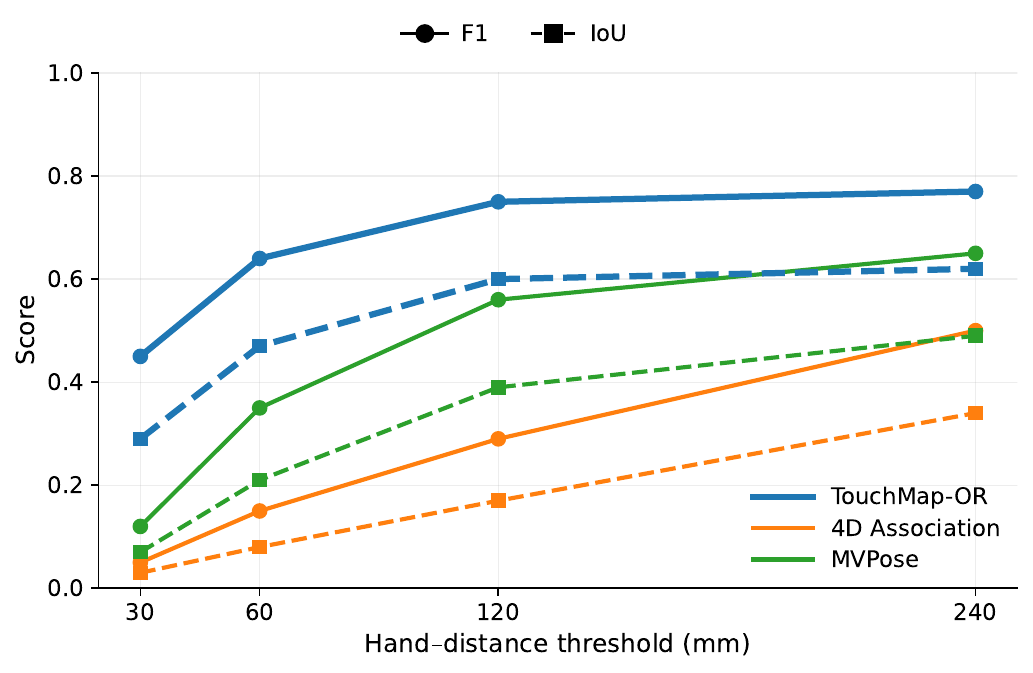}
  \caption{Binary contact detection performance as a function of the hand–surface distance threshold. We report F1-score and IoU computed from framewise contact predictions across the dataset.}
  \label{fig:overview}
\end{figure}

Table~\ref{tab:tracking_results} reports multi-person tracking performance for the evaluated methods. MVPose achieves the highest IDF1 score (0.49), while TouchMap-OR reaches a comparable IDF1 of 0.48. However, TouchMap-OR substantially improves identity stability, reducing the number of ID switches from 236 for MVPose and 547 for 4D Association to only 108. This indicates that while overall association accuracy is similar, the proposed pipeline produces significantly more temporally consistent identity tracks, which is important for reliable reasoning about interactions over time. Table~\ref{tab:contact_identity_results} shows the performance of the methods on the core task of hand–surface interaction analysis. TouchMap-OR substantially outperforms the baselines across all contact metrics. Our method achieves a binary contact F1-score of 0.75 and an IoU of 0.60, improving over MVPose-based heuristics (0.56 F1) and 4D Association (0.29 F1). Semantic contact recognition also improves to an F1-score of 0.37, indicating more accurate attribution of contacts to specific surfaces. Episode recall increases to 0.54, demonstrating that more than half of the annotated interaction episodes are successfully detected. Finally, TouchMap-OR achieves the highest identity attribution accuracy (0.96), confirming that the integration of articulated hand reconstruction with scene semantics enables reliable identification of which clinician interacts with which surface.

%% file: sec/6_discussion.tex
\section{Discussion}
Our experiments demonstrate that TouchMap-OR enables reliable reasoning about hand–surface interactions in complex operating room environments. While the multi-person tracking performance is slightly lower than specialized tracking baselines (Table~\ref{tab:tracking_results}), the obtained identity consistency is sufficient to support downstream interaction analysis. In contrast, TouchMap-OR substantially improves the core task of hand–surface interaction detection and attribution (Table~\ref{tab:contact_identity_results}), achieving markedly higher contact detection and semantic interaction recognition compared to heuristic approaches built on existing tracking pipelines. This suggests that accurate modeling of articulated hands together with a semantically structured 3D scene is more important for interaction understanding than maximizing raw tracking metrics alone. The gap between binary and semantic contact F1 reflects a characteristic challenge of the operating-room environment: many surfaces often occur in close proximity, such as bandages, venous catheters, or blankets. In such situations geometric proximity may correctly detect a contact event but assign it to the wrong surface label. The higher binary performance therefore indicates reliable interaction detection, while the lower semantic score highlights the difficulty of precise object attribution in cluttered clinical scenes and the limitations of purely geometric contact inference. Tracking remained stable under typical operating-room conditions such as occlusions, motion, and close proximity between clinicians. However, the 4D Association tracker occasionally produced erroneous tracks for the patient, whose skeleton could become entangled with nearby clinicians. We attempted to mitigate this with a simple spatial filter removing tracks overlapping the patient point cloud, but the issue persisted in some sequences, highlighting limitations of general-purpose tracking methods in clinical environments where a largely static patient occupies the scene center. Finally, our semantic 3D reconstruction enables reasoning over both fixed surfaces and movable or deformable objects such as oxygen tubes and tourniquets. However, thin or transparent items—including IV lines, shavers, or peripheral venous catheters—were often incompletely segmented, resulting in partial or noisy reconstructions. Suspended cables were particularly challenging, as background leakage frequently distorted their geometry.



%

%% file: sec/7_conclusion.tex
\section{Conclusion}
We presented TouchMap-OR, a multi-camera computer vision system for estimating HSC of healthcare workers in real operating rooms. Our results demonstrate stable identity tracking and reliable contact detection across diverse object categories, including both deformable and movable equipment. Beyond its technical performance, TouchMap-OR offers a scalable foundation for quantifying hand-associated transmission networks in clinical environments. Such analyses can support data-driven infection prevention strategies and open new perspectives on the dynamics of human–environment interaction in healthcare. 

%% file: sec/8_declarations.tex
\section*{Declarations}

\noindent\textbf{Conflict of Interest} 
The authors declare that they have no competing interests.

\noindent\textbf{Ethical Approval} 
The study protocol (Req-2023-00636) was submitted to the Cantonal Ethics Commission of Zurich, which confirmed that the project does not fall under the scope of the Swiss Human Research Act and therefore does not require formal ethics approval. 

\noindent\textbf{Informed Consent} Written informed consent was obtained from all participating patients. Healthcare staff were verbally informed about the study procedures prior to data collection and were given the option to decline participation.